\documentclass[journal]{IEEEtran}

\usepackage{amsmath}
\usepackage{mathtools}
\usepackage{caption}
\usepackage{subcaption}
\usepackage{cite}

\ifCLASSINFOpdf
\else
\fi
\begin{document}

\title{Towards On-Board Panoptic Segmentation of Multispectral Satellite Images}
%
%
%

\author{Tharindu~Fernando,~\IEEEmembership{Member,~IEEE,} 
        Clinton~Fookes,~\IEEEmembership{Senior Member,~IEEE,}
        Harshala~Gammulle,~\IEEEmembership{Member,~IEEE,} 
        Simon~Denman,~\IEEEmembership{Member,~IEEE,}
        ~and~ Sridha~Sridharan,~\IEEEmembership{Life Senior Member,~IEEE.}

\IEEEcompsocitemizethanks{\IEEEcompsocthanksitem T. Fernando, C. Fookes, H. Gammulle, S.Denman, and S. Sridharan  are with The Signal Processing, Artificial Intelligence and Vision Technologies (SAIVT), Queensland University of Technology, Australia.\protect }}

%
%

\markboth{Journal of \LaTeX\ Class Files,~Vol.~14, No.~8, August~2015}%
{Shell \MakeLowercase{\textit{et al.}}: Bare Demo of IEEEtran.cls for IEEE Journals}
%



\maketitle

\begin{abstract}
With tremendous advancements in low-power embedded computing devices and remote sensing instruments, the traditional satellite image processing pipeline which includes an expensive data transfer step prior to processing data on the ground is being replaced by on-board processing of captured data. This paradigm shift enables critical and time-sensitive analytic intelligence to be acquired in a timely manner onboard the satellite itself. However, at present, the on-board processing of multi-spectral satellite images is limited to classification and segmentation tasks. Extending this processing to it's next logical level, in this paper we propose a lightweight pipeline for on-board panoptic segmentation of multi-spectral satellite images. Panoptic segmentation offers major economic and environmental insights, ranging from yield estimation from agricultural lands to intelligence for complex military applications. Nevertheless, the on-board intelligence extraction raises several challenges due to the loss of temporal observations and the need to generate predictions from a single image sample. To address this challenge, we propose a multimodal teacher network based on a cross-modality attention-based fusion strategy to improve the segmentation accuracy by exploiting data from multiple modes. We also propose an online knowledge distillation framework to transfer the knowledge learned by this multi-modal teacher network to a uni-modal student which receives only a single frame input, and is more appropriate for an on-board environment. We benchmark our approach against existing state-of-the-art panoptic segmentation models using the PASTIS multi-spectral panoptic
segmentation dataset considering an on-board processing setting. Our evaluations demonstrate a substantial 10.7\%, 11.9\% and 10.6\% increase in Segmentation Quality (SQ), Recognition Quality (RQ), and Panoptic Quality (PQ) metrics compared to the existing state-of-the-art model when it is evaluated in an on-board processing setting.
\end{abstract}


\IEEEpeerreviewmaketitle

\section{Introduction}

\IEEEPARstart{F}{ollowing} the launch of Landsat-1, the first satellite to carry a multi-spectral remote sensing instrument, multi-spectral imaging of earth from space has evolved rapidly. The availability of additional spectral bands via multi-spectral remote sensing enables better segmentation and differentiation of objects compared to using the visible spectrum alone. 

The traditional image analysis process for satellite image processing contains typical stages of: (i) data acquisition, (ii) data transfer from satellite to ground, (iii) centralized ground processing, and (iv) information extraction \cite{xu2021board}. This process is slow, in particular due to the time required to down-link a high resolution multi-band image, meaning that extracted intelligence may be antiquated for effective decision making. As such, a new paradigm of algorithms are being introduced \cite{qiu2021novel, xu2021board} which can process captured satellite images on-board. This also avoids the expensive and time-consuming data transmission of the original data, and the limited transmission bandwidth can be used to convey the extracted intelligence. The rapid advancements in low-power embedded Graphic Processing Units (GPUs), such as the NVIDIA Jetson series, has further advanced this line of work \cite{xu2021board, smartsat2022}. In line with this evolution, we propose to take the on-board processing of satellite acquired multi-spectral images to its the next logical level. Expanding from classification \cite{liu2021novel, gao2021hyperspectral} and segmentation tasks \cite{jung2021boundary, hong2020multimodal}, we seek to perform panoptic segmentation using limited on-board resources.   

Since its first introduction by Kirillov et. al \cite{kirillov2019panoptic}, panoptic segmentation has gained substantial traction in numerous computer vision applications. Specifically, panoptic segmentation has helped to eliminate the dichotomy between the recognition of countable objects such as people and animals, and the segregation of uncountable \textit{stuff}, such as sky and roads, by proposing a unified framework to achieve both. Panoptic segmentation is essentially a multi-task architecture which jointly performs semantic segmentation and object detection (or instance segmentation), where the former is used for recognising uncountable, while the latter is employed to segregate countable objects.

Panoptic segmentation enables unmet analytic capabilities in remote sensing, ranging from monitoring urban areas and agricultural lands to military applications \cite{de2022panoptic}. For instance, while semantic segmentation outputs can be used to obtain a general overview of the yield from individual crop types, instance segmentation of agricultural parcels offers an in depth analysis regarding the yield from individual parcels. Similarly, in urban monitoring, advanced intelligence regarding individual households or road/traffic conditions can be obtained by using panoptic segmentation and considering each detected instance. 

\begin{figure*}
    \centering
    \includegraphics[width=.85 \textwidth]{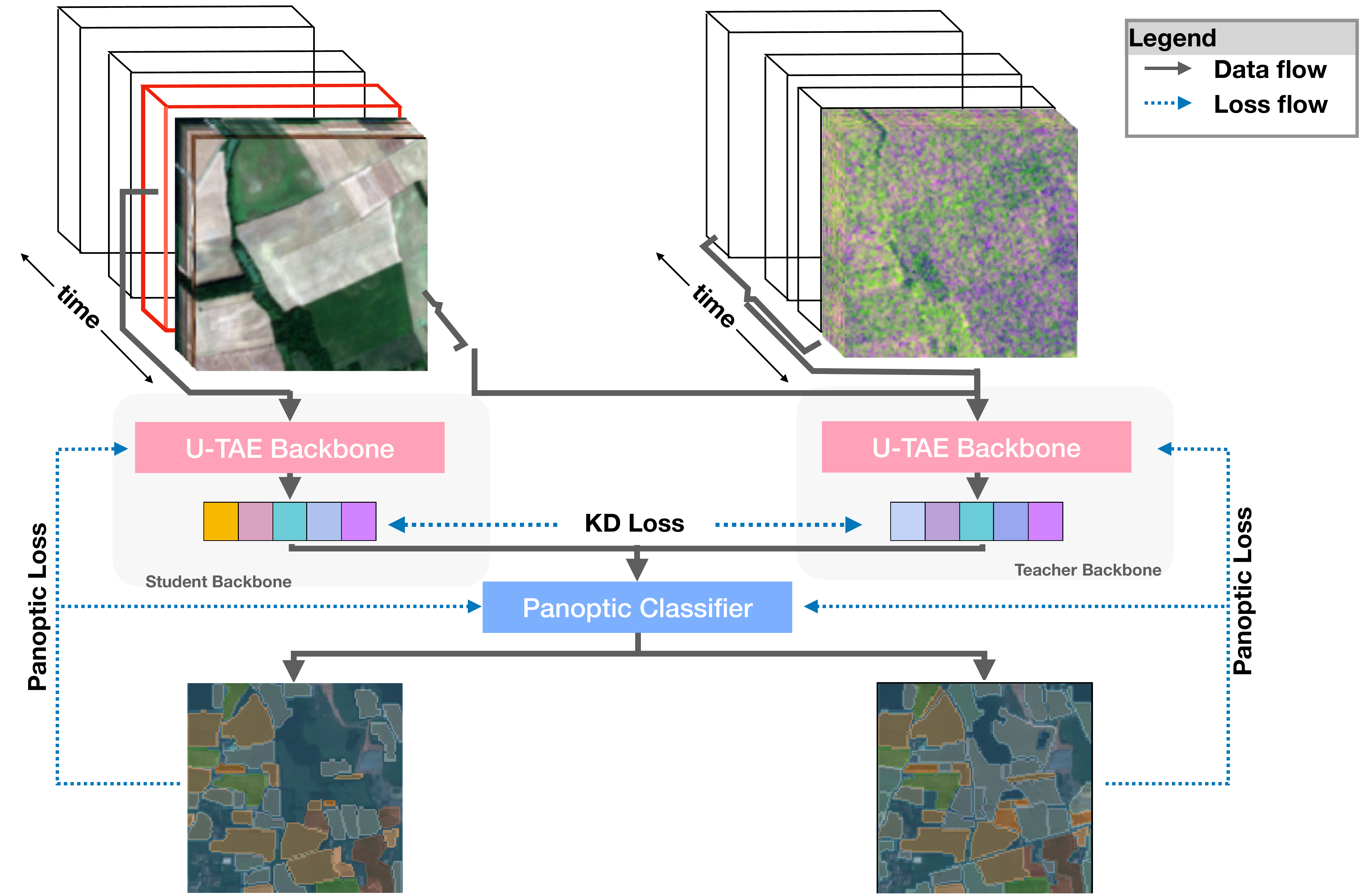}
    \caption{The proposed cross modality knowledge distillation pipeline. We use a multi-modal teacher that receives a time-series input to guide a light-weight, uni-modal, single frame student network.}
    \label{fig:model}
\end{figure*}

In this paper, a light-weight framework for multi-spectral panoptic segmentation is proposed. Specifically, our contributions are three-fold. First, we benchmark the performance of several state-of-the-art panoptic segmentation models in the on-board processing setting using the Panoptic Agricultural Satellite TImeSeries (PASTIS) \cite{garnot2021panoptic} dataset. To the best of our knowledge, this is the only publicly available multi-spectral dataset with panoptic annotation. Second, we propose a multi-modal fusion architecture to improve panoptic segmentation performance in the on-board computation setting. Third, we propose a novel cross modality knowledge distillation pipeline to transfer the knowledge from this aforementioned multi-modal teacher network to a light-weight uni-modal student, such that it achieves superior performance. This framework is visually illustrated in Fig. \ref{fig:model}. 

The rest of the paper is organised as follows. In Sec. \ref{sec:related_works} we summarise recent developments in related areas and illustrate how the proposed work deviates from these. Sec. \ref{sec:materials_Methods} describes the details of the datasets used, the architectures of our multi-modal and uni-modal networks, and the knowledge distillation pipeline. Sec. \ref{sec:evaluations} presents the experimental results and conclusions are drawn in Sec. \ref{sec:conclusions}. 

\section{Related Works}
\label{sec:related_works}

\subsection{On-board Processing of Satellite Images}

Since the launch of the quickbird satellite with high resolution cameras and high capacity on-board processing capabilities, there has been an increase in research and development related to advanced on-board satellite image processing. Numerous works have investigated on-board capabilities of object detection \cite{arechiga2018onboard, xu2021board}, image classification \cite{gretok2021onboard}, image selection \cite{wang2015onboard}, denoising and artefact removal \cite{cucchetti2021onboard}, and change detection \cite{sophiayati2009onboard, yuhaniz2005embedded}.

Specifically, considering the most relevant literature in the domains of on-board object detection and classification, in \cite{arechiga2018onboard} the authors benchmark a light-weight CNN based classifier for ship detection using an Nvidia Jetson TX2 GPU. The authors demonstrated the utility of such a CNN architecture over traditional machine learning techniques. More recently, A You Only Look Once v4 (YOLOv4)-Tiny \cite{bochkovskiy2020yolov4} architecture has been benchmarked by \cite{xu2021board} for ship detection using HISEA-1 SAR images. The Constant False Alarm Rate (CFAR) algorithm is used to first extract target regions for deep learning-based ship detection. This significantly reduces the number of image patches submitted to the computationally complex deep learning based ship detection algorithm. The detected targets from the CFAR algorithm are stored as 256 $\times$ 256 pixel patches which are fed to the YOLOv4-Tiny object detector. Finally, the outputs of the detector are mapped to the original SAR image coordinates using bilinear interpolation. As such, only relevant intelligence is transferred via the satellite-earth data transmission system. In a similar line of work, \cite{gretok2021onboard} applied transfer learning to adapt a MobileNetV2 \cite{sandler2018mobilenetv2} classifier pretrained on ImageNet \cite{deng2009imagenet} for the classification of earth-observation imagery at different ground-resolved distances. However, to the best of our knowledge, none of the existing works have investigated the use of panoptic segmentation models, which are highly computationally complex in comparison to object detectors explored to date in on-board processing scenarios. 

A complementary line of research \cite{adams2021hardware, goodwill2020adaptively, hihara2015onboard, bruhn2020enabling} has focused on hardware and software acceleration techniques to enable efficient on-board processing of satellite images. However, these methods optimise matrix multiplication operations and standard convolution and other neural network architecture computations, enabling operation over compressed inputs, and do not propose a specific architecture for the on-board computation. Therefore, we do not review these works in detail, however, acknowledge the contributions of these works towards enabling efficient on-board AI capabilities. 

\subsection{Knowledge Distillation}

Cross modality knowledge distillation has recently gained attention, but there are limited works concerning the transfer of knowledge from a multi-modal teacher to a uni-modal student \cite{gou2021knowledge}. Specifically, Garcia et. al \cite{garcia2018modality} has used depth and RGB images to train a multi-modal teacher while Gao et. al \cite{gao2019privileged} has used multi-modal medical images as the input to the teacher. Furthermore, to the best of our knowledge, none of the existing works in cross-modality distillation have used on-line distillation. 

In on-line distillation, both teacher and student networks are simultaneously updated, creating an end-to-end trainable pipeline. However, such mutual learning (or joint learning) in a high capacity environment is an open research question which requires further exploration and careful consideration of the relationships between the student and the teacher \cite{gou2021knowledge}. For instance, recently an online knowledge distillation framework that uses an ensemble of students and teacher networks is proposed in \cite{guo2020online}, and these networks are collaboratively trained. Chen et. al \cite{chen2020online} have extended this concept to use auxiliary peers and group leaders such that more supervision is provided to the student and diverse knowledge distillation can occur. However, these application settings are quite different to the proposed on-board multispectral panoptic segmentation setting which has a unique set of challenges and requirements. Using multiple student networks is not feasible due to on-board hardware requirements. As such, by leveraging the multiple modalities that are available for the teacher network we propose to overcome the challenges with our application due to the loss of temporal information.  

\section{Materials and Methods}
\label{sec:materials_Methods}

\subsection{Datasets}
\subsubsection{Panoptic Agricultural Satellite TImeSeries (PASTIS) \cite{garnot2021panoptic} dataset}

The PASTIS dataset is composed of 2,433 multi-spectral image patches, each with 10 spectral bands and of size 128 $\times$ 128 pixels, at 10m/pixel resolution. To the best of our knowledge, this is the only publicly available multi-spectral satellite image dataset with panoptic annotation. This data has been captured using the Sentinel-2 platform and four Sentinel tiles have been used. The authors have generated a time-series dataset by stacking all the available observations for each patch between September 2018 and November 2019. 

However, the temporal sampling of the data is irregular as the orbital schedule of Sentinel-2 means patches are observed a different numbers of times and at different intervals. Furthermore, data providers have not processed the images if there is more than 90\% cloud coverage, which leads to irregular length time-series. The temporal length of the observations ranges from 33 to 61 samples per sequence. 

The dataset authors have used the publicly available French Land Parcel Identification System to retrieve the extents and the content of each parcel when annotating the patches. Within the annotations, each pixel has a semantic label that represents the crop type or the background class. Non-agricultural land is classified as the background class. In total there are 18 crop types, the background class and a void class, resulting in 20 semantic classes. The void class denotes agricultural parcels that contain crops that are not classified. 
The authors have also proposed a 5-fold cross validation scheme that divides the 2,433 patches into 5 splits. When splitting the data, care was taken to ensure that each fold has patches from all 4 Sentinel tiles and each split contains comparable class distributions. Furthermore, cross-contamination of data caused by adjacent patches falling into different splits was avoided. 

\begin{figure*}
\centering
\begin{subfigure}{0.23\textwidth}
    \includegraphics[width=\textwidth]{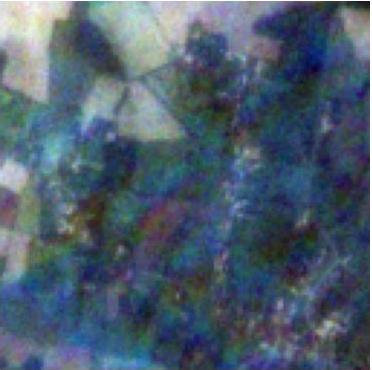}
    \caption{Frame 4}
\end{subfigure}
\hfill
\begin{subfigure}{0.23\textwidth}
    \includegraphics[width=\textwidth]{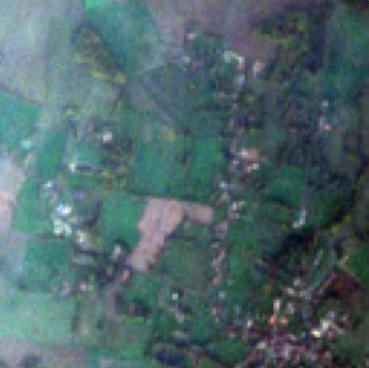}
    \caption{Frame 7}
\end{subfigure}
\hfill
\begin{subfigure}{0.23\textwidth}
    \includegraphics[width=\textwidth]{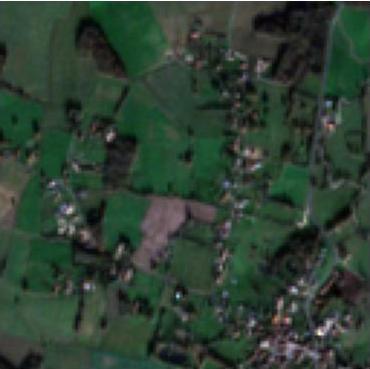}
    \caption{Frame 8}
\end{subfigure}
\hfill
\begin{subfigure}{0.23\textwidth}
    \includegraphics[width=\textwidth]{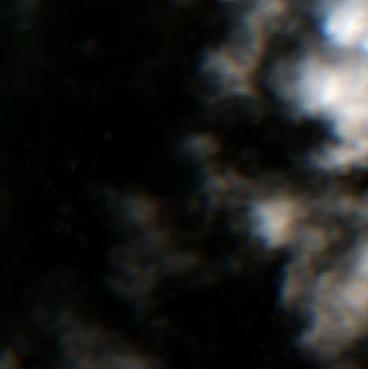}
    \caption{Frame 9}
\end{subfigure}

\begin{subfigure}{0.23\textwidth}
    \includegraphics[width=\textwidth]{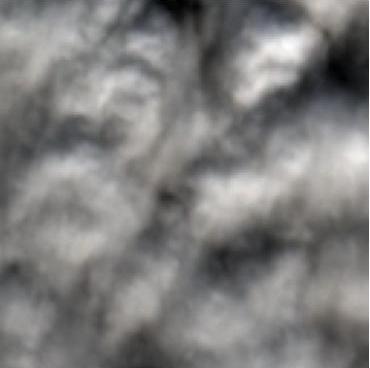}
    \caption{Frame 13}
\end{subfigure}
\hfill
\begin{subfigure}{0.23\textwidth}
    \includegraphics[width=\textwidth]{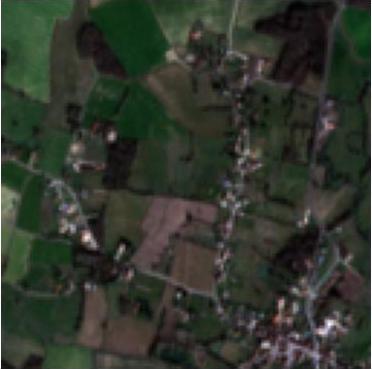}
    \caption{Frame 16}
\end{subfigure}
\hfill
\begin{subfigure}{0.23\textwidth}
    \includegraphics[width=\textwidth]{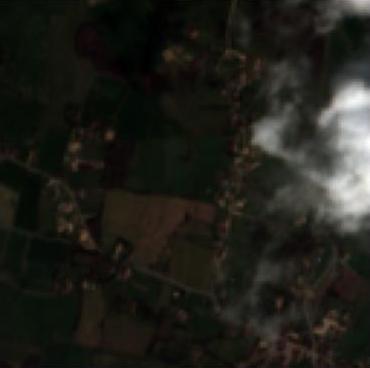}
    \caption{Frame 17}
\end{subfigure}
\hfill
\begin{subfigure}{0.23\textwidth}
    \includegraphics[width=\textwidth]{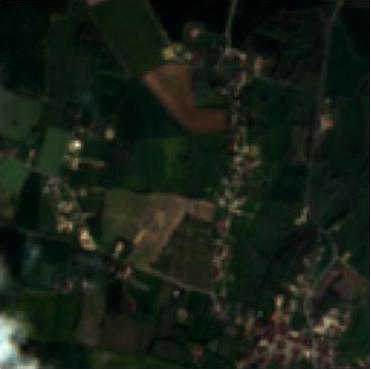}
    \caption{Frame 20}
\end{subfigure}
        
\caption{Visualisations of sample frames from the S2\_30634 time-series of the PASTIS \cite{garnot2021panoptic} dataset. The challenges due to cloud cover, illumination conditions and other atmospheric noise is clearly visible in this sequence.}
\label{fig:single_frame_challenges}
\end{figure*}

In \cite{garnot2021panoptic} the authors have observed that even after filtering out predominantly cloudy acquisitions, there still exists partially or completely obstructed patches. Furthermore, they highlight the utility of time-series data for resolving those obstructions and boundary ambiguities due to clouds and other atmospheric noise. For instance, in Fig. \ref{fig:single_frame_challenges} we visualise the image sequence of a randomly chosen patch. Within the frames it can be seen that there exists boundary ambiguities which are indistinguishable even for humans. The multiple temporal observations can be used to eradicate these ambiguities in ground-based processing, however, in an on-board setting an AI algorithm does not have access to such historical data. In particular, when benchmarking the algorithms presented in \cite{garnot2021panoptic} in a non-temporal (single frame) setting we illustrate how state-of-the-art methods struggle to generate accurate predictions (See Sec. \ref{sec:benchmarking}). Hence we propose an on-line knowledge distillation strategy to augment a light-weight model that observes only a randomly chosen single frame of the time-series, allowing it to \textit{imagine} the corresponding features from the time-series and achieve more accurate panoptic segmentation. 

\subsubsection{PASTIS-R \cite{sainte2021multi} dataset}

PASTIS-R is an extension of the PASTIS dataset where the authors have combined the Sentinel-2 multi-spectral patches with the corresponding Sentinel-1 observations. Specifically, they have used Sentinel-1 data in ground range detected format and assembled these into 3-channel images (vertical polarization, horizontal polarization and the ratio of vertical over horizontal polarization). Separate observations were captured in both ascending and descending orbits, generating two 3-channel images for each observation. This multi-modal dataset has rich spectral information coming from the Sentinel-2 multi-spectral observations, while the C-band radar of Sentinel-1 captures useful geometric information as it's immune to cloud cover \cite{sainte2021multi}.

Hence, this multi-modal dataset can be seen as containing complementary information to be leveraged when views are obstructed due to cloud and other illumination conditions, however, it is illogical to directly fuse the data in an on-board setting as data is captured from separate satellites. Therefore, we propose a knowledge distillation strategy where a heavy-weight multi-modal teacher guides a uni-modal single frame student such that it can imagine the unseen features which are only visible to the teacher.

\subsection{Knowledge Distillation Framework}

In this section we introduce the structure of the multi-modal backbone that we use for our teacher network (Sec. \ref{sec:teacher}) and the architecture of the light-weight uni-modal backbone of the student network (Sec. \ref{sec:student_backbone}). In Sec. \ref{sec:Panoptic_Classifier} we describe how the extracted features from these backbones are used in the panoptic classifier while Sec. \ref{sec:loss_fuctions} outlines the loss functions used to train the network. 

\subsubsection{A Teacher Backbone with Cross-Modality Attention}
\label{sec:teacher}
In this section we illustrate how the proposed cross-modality attention backbone is formulated. The approach takes inspiration from the U-TAE backbone from \cite{garnot2021panoptic}, though we incorporate specific augmentations to increase its capabilities in a multi-modal setting. The spatio-temporal encoding structure is based on \cite{garnot2021panoptic}, however, prior to the decoding phase we employ cross-modality attention to extract task specific, salient information that compliments each modality. 

The inputs to the teacher network, $\dot{X}$ and $\ddot{X}$, are spatio-temporal inputs with shapes $T \times \dot{C} \times W \times H$ and $T \times \ddot{C} \times W \times H$, respectively, where $T$ denotes the length of the time-series (i.e number of frames), $W$ indicates the width of the image and $H$ denotes its height. $\dot{C}$ and $\ddot{C}$ represents the number of channels in the multi-spectral and radar modalities, respectively. Each spectral encoder, $\dot{f}_e^l$ and $\ddot{f}_e^l$ ($l \in [1, \ldots, L]$), has $L$ levels where each level is comprised of convolutions, Relu and normalisation layers. The encoder $f_e^l$ ($f_e^l \in [\dot{f}_e^l$, $\ddot{f}_e^l]$) at level $l$ receives the output $\dot{e}^{l-1}_t$ ($t \in [1, \ldots, T]$) of the immediately proceeding layer and generates a feature map $\dot{e}^{l}_t$. At each level the encoder maps are compressed using strided convolutions such that the output height $H^l = \frac{H}{2^{l-1}}$, and width $W^l = \frac{W}{2^{l-1}}$. All frames in the sequence are simultaneously processed by the encoders. 

To perform temporal encoding, similar to \cite{garnot2021panoptic}, we use a Light-weight Temporal Attention Encoder (L-TAE) \cite{garnot2020lightweight} which is a multi-head self-attention network with $G$ heads. The feature maps $\dot{e}^l$ and $\ddot{e}^l$ of the last level (i.e $L$) are used and the L-TAE networks, $\dot{f}_{L-TAE}$ and $\ddot{f}_{L-TAE}$ generate $G$ attention vectors, each of shape $T \times W_L \times H_L$ as,
\begin{equation}
\begin{split}
[\dot{a}^{L,1}, \ldots, \dot{a}^{L,G}] = \dot{f}_\mathrm{L-TAE}(\dot{e}^l), \\
[\ddot{a}^{L,1}, \ldots, \dddot{a}^{L,G}] = \dot{f}_\mathrm{L-TAE}(\ddot{e}^l).
\end{split}
\label{eq:l-tae_att}
\end{equation}

To apply these attention maps to all encoded maps from the U-Net, interpolation is performed, where each attention vector, $\dot{a}^{l,g}$ and $\ddot{a}^{l,g}$ in $g$ is interpolated to size $H_l$ and $W_l$ for level $l \in [1, \ldots, L]$. The the respective channels $\dot{C}$ and $\ddot{C}$ of the multi-spectral and radar modalities are split into $G$ consecutive groups (i.e. $[\dot{e}^{l,1}, \ldots, \dot{e}^{l,G}]$, each of shape $T \times \frac{\dot{C}_l}G \times W_l \times H_l$; and $[\ddot{e}^{l,1}, \ldots, \ddot{e}^{l,G}]$, each of shape $T \times \frac{\ddot{C}_l}G \times W_l \times H_l$). 

Then we multiply each embedding by it's respective attention map, average the temporal sequence, and pass the resultant feature vector through a $1 \times 1$ convolution layer,
\begin{equation}
\begin{split}
    \dot{h}^l = \dot{f}^l_\mathrm{Conv_{1 \times 1}} \left[\sum_{t=1}^T \dot{a}^{l, g}_t \dot{e}^{l, g}_t\right]_{g=1}^G, \\
    \ddot{h}^l = \ddot{f}^l_\mathrm{Conv_{1 \times 1}} \left[\sum_{t=1}^T \ddot{a}^{l, g}_t \ddot{e}^{l, g}_t\right]_{g=1}^G, 
\end{split}
\end{equation}
where $[.]$ denotes concatenation along the channel dimension.

In \cite{sainte2021multi}, the authors have introduced several fusion strategies for the PASTIS-R dataset. Their evaluations using early, late, decision, and mid level fusion indicated that early fusion is the best strategy for panoptic segmentation. However, leveraging the generated attention vectors, $\dot{a}^{l,g}$ and $\ddot{a}^{l,g}$, we propose fusion of the multi-spectral and radar modalities using cross-modality attention. Specifically, we generate attention maps, $\dot{a}^{l}$ and $\ddot{a}^{l}$ for each modality such that,
\begin{equation}
\begin{split}
    \dot{a}^l = \frac{\left[\sum_{t=1}^T\sum_{g=1}^G \dot{a}^{l,g}_t \right]}{T \times G}, \\
    \ddot{a}^l = \frac{\left[\sum_{t=1}^T\sum_{g=1}^G \ddot{a}^{l,g}_t \right]}{T \times G}.
\end{split}
\end{equation}

These attention maps denote how salient features are spatially arranged in a particular modality. Hence, they are important for capturing complimentary information from a second mode. We interpolate these 2D maps to match the channel dimension of the complementary mode, and generate an augmented feature such that,
\begin{equation}
    \hat{h}^l = {f}^l_\mathrm{Conv_{1 \times 1}} \left[ \dot{h}^l, \ddot{h}^l, \dot{h}^l\ddot{a}^l, \ddot{h}^l\dot{a}^l \right],
\end{equation}
where $[,]$ denotes concatenation in the channel dimension. Then the augmented feature $\hat{h}^l$ is used by the spatial decoder. 

Following the conventional U-Net network structure, the encoder map from level $l$, $\hat{h}^l$, is concatenated with the decoded feature map, $\hat{d}^{l-1}$, from level $l-1$. Each decoder block is composed of strided transposed convolutions to up sample the feature maps from the previous level. The decoder output is formally denoted as $\hat{d} = [\hat{d}^1, \ldots, \hat{d}^L]$.

\subsubsection{A Uni-Modal Single Frame Backbone for the Student}
\label{sec:student_backbone}

The student backbone receives the input, $\Tilde{X}$, of shape $\dot{C} \times W \times H$, that corresponds to a single frame from the multi-spectral modality. Similar to the teacher backbone we pass it through a spatial encoder $\Tilde{f}_e^l$ with $L$ levels and generate the feature map $e^l$. However, in contrast to the teacher backbone, $\Tilde{e}^l$ is of dimension $C_l \times W_l \times H_l$. Despite the absence of the temporal dimension, the L-TAE is used in the student backbone with the motivation of capturing the channel-wise relationships with the aid of a multi-head attention formulation such that,
\begin{equation}
    [\Tilde{a}^{L,1}, \ldots, \Tilde{a}^{L,G}] = \Tilde{f}_\mathrm{L-TAE} (\Tilde{e}^L),
\end{equation}
however, when applying the attention no temporal summarisation is performed. Specifically,
\begin{equation}
    \Tilde{h}^l = \Tilde{f}^l_\mathrm{Conv_{1 \times 1}} \left[\Tilde{a}^{l, g} \Tilde{e}^{l, g}_t\right]_{g=1}^G,
\end{equation}
where $[.]$ denotes concatenation along the channel dimension. Then, similar to the teacher backbone, the spatial decoder of the student up samples these features to generate the multi-scale output features, $\Tilde{d} = [\Tilde{d}^1, \ldots, \Tilde{d}^L]$.

\subsubsection{Panoptic Classifier}
\label{sec:Panoptic_Classifier}

The panoptic classifier receives multi-scale feature maps (either from the teacher or student backbones) and produces predictions over parcels. We used the same architecture as the panoptic classifier network from \cite{garnot2021panoptic}, where it makes four predictions for each parcel: a center point, a bounding box, a binary instance mask and a class. We refer the readers to \cite{garnot2021panoptic} for further details regarding this network's architecture. 

We use a single instance of this panoptic classifier to evaluate the predictions of both student and teacher backbones, allowing it to identify the similarities and correspondences between the feature vectors $\hat{d}$ and $\Tilde{d}$. Note that in our formulation, the dimensions of the decoded maps from the student and teacher backbones at each level of the U-Net are identical.

\subsubsection{Loss Functions for On-line Knowledge Distillation}
\label{sec:loss_fuctions}

As discussed in Sec. \ref{sec:Panoptic_Classifier}, the panoptic classifier makes four predictions, a centerness heat map, bounding boxes, instance masks and classes, and as such \cite{garnot2021panoptic} used four losses to supervise these predictions. 

Formally, let $\breve{m} \in [0, 1]^{W \times H}$ denote the ground truth centerness heat map, let $m$ be the predicted centerness heat map, and let there be $P$ ground truth parcels. Then $L_\mathrm{center}$ is defined as,
\begin{equation*}
\resizebox{\hsize}{!}{$
L_\mathrm{center} = \frac{-1}{|P|} \sum_{\substack{i=1 \ldots H \\ j=1 \ldots W}} = \begin{cases}
log (m_{i, j}) & \text{if } \breve{m}_{i,j}=1,\\
(1 - \breve{m}_{i,j})^\beta log(1-m_{i, j}) & \text{else,}
\end{cases}$}
\end{equation*}
where $\beta$ is a hyper-parameter and following \cite{garnot2021panoptic} it is set to 4.

Then $P'$ is defined as a set of detected parcels and $k'(p)$ denotes the predicted class of the parcel $p \in P'$, while $k''(p)$ indicates its ground truth class. Then the class loss of parcel $p$ is defined as,
\begin{equation}
    L^p_\mathrm{class} = -log(k'(p) [k''(p)]).
\end{equation}

Let $h'(p)$ and $h''(p)$ denote the predicted and ground truth heights of the bounding boxes of parcel $p$, and $w'(p)$ and $w''(p)$ indicate its respective width. Then,
\begin{equation}
    L^p_\mathrm{size} = \frac{|h'(p) - h''(p)|}{h''(p)} + \frac{|w'(p) - w''(p)|}{w''(p)},
\end{equation}
defines the size loss. In addition a pixel-wise Binary Cross Entropy (BCE) loss is computed between the predicted shape, $l'(p)$ and the corresponding ground truth instance $s''(p)$ mask when it is cropped using the predicted bounding box. This can be defined as,
\begin{equation}
    L^p_\mathrm{shape} = \mathrm{BCE} (l'(p), \mathrm{crop(s''(p))}).
\end{equation}

Then the complete loss function in \cite{garnot2021panoptic} can be written as,
\begin{equation}
    L = L_\mathrm{centre} + \frac{1}{P'} \sum_{p\in P'}(L^p_\mathrm{class} + L^p_\mathrm{size} + L^p_\mathrm{shape}).
\label{eq:baseline_loss}
\end{equation}

To facilitate an on-line student-teacher knowledge distillation framework we perform the following augmentations. Let $f_\mathrm{panoptic}$ denote the output of the panoptic classifier introduced in Sec. \ref{sec:Panoptic_Classifier} and let $L$ denote the complete loss defined in Eq. \ref{eq:baseline_loss}. Then,
\begin{equation}
    \begin{split}
        L_\mathrm{student} = L(f_\mathrm{panoptic} (\{\Tilde{d}\}_1^L)), \\
        L_\mathrm{teacher} = L(f_\mathrm{panoptic} (\{\hat{d}\}_1^L)), \\
        L_\mathrm{distil} = \sum_{i=1}^L ||\Tilde{d}_i -[\hat{d}_i]^{\mathrm{detach}} ||^2,
    \end{split}
\end{equation}
where $[\hat{d}_i]^{\mathrm{detach}}$ denotes detaching the tensor from gradient computation. Then the final loss of our framework is defined as,
\begin{equation}
L^* = L_\mathrm{student} + \lambda_1L_\mathrm{teacher} + \lambda_2L_\mathrm{distil},
\end{equation}
where $\lambda_1$ and $\lambda_2$ are hyper-parameters controlling the contributions of the respective loss terms. 

\section{Evaluations}
\label{sec:evaluations}

In Sec. \ref{sec:fusion} we first report the performance of the proposed fusion architecture when the time-series PASTIS-R dataset's input is used. In Sec. \ref{sec:benchmarking} evaluations of the state-of-the-art panoptic segmentation methods in the on-board setting without time-series data (i.e. single input frame) are provided. In Sec. \ref{sec:Knowledge_Distillation_results}, we evaluate the proposed knowledge distillation pipeline, and in Sec. \ref{sec:qualitative_results} we qualitatively illustrate how the student features have been augmented by the teacher. 

Similar to \cite{kirillov2019panoptic} we report class averaged Segmentation Quality (SQ), Recognition Quality (RQ) and Panoptic Quality (PQ), where RQ corresponds to the quality of the detection and can be computed using,
\begin{equation}
    RQ = \frac{|TP|}{|TP| + \frac{1}{2}|FP| + \frac{1}{2}|FN|},
\end{equation}
where TP, FP, and FN denotes true positive, false positive and false negative, respectively. SQ is computed as the average IoU of matched segments and can be written as,
\begin{equation}
    SQ = \frac{\sum_{(i,j) \in TP} IoU(i,j)}{|TP|},
\end{equation}
where $i,j$ denotes the detected parcels within the set of true positives. Then PQ is the product of SQ and RQ (i.e. $PR = SQ \times RQ$), thus combining the detection, classification and delineation accuracies \cite{sainte2021multi}. 

\subsection{Results of the Fusion Model}
\label{sec:fusion}
In Tab. \ref{tab:fusion_results} we report the performance of the proposed cross attention based fusion model together with the trainable parameter counts and run times. Note that the run times are reported for evaluating a single sample with 40 frames in the time-series. In addition, we report the performance of the existing baselines and attention variations for comparison. Note that due to sampling rate inconsistencies between the modalities, similar to \cite{sainte2021multi}, we have interpolated Radar observations to match the time stamps of the multi-spectral modality. 

\begin{table}[htbp]
\centering
\resizebox{\linewidth}{!}{%
\begin{tabular}{|c|c|c|c|c|c|}
\hline
Model                 & SQ & RQ & PQ & Parameters & Run Time \\ \hline
U-TAE+PAPS-Multispec \cite{garnot2021panoptic}   & 81.3   & 49.2   & 40.4   & 1.2M            &   0.53s       \\ \hline
U-TAE+PAPS-Radar     & 77.2   & 39.1   & 30.8   & 1.2M            &  0.38s        \\ \hline
U-TAE+PAPS-Late Fusion \cite{sainte2022multi}          & 81.6   & 50.5   & 41.6   & 2.3M           & 0.73s         \\ \hline
U-TAE+PAPS-Early Fusion \cite{sainte2022multi}         & 82.2   & 50.6   & 42.0   & 1.7M           & 0.52s          \\ \hline
Self Attention Fusion & 82.3   & 53.1   & 43.8   &  2.4M          & 0.89s         \\ \hline
Cross Attention Fusion & \textbf{82.7}    & \textbf{55.6}    & \textbf{45.8}    &  2.4M          &  0.98s         \\ \hline
\end{tabular}}
\vspace{2px}
\caption{Evaluation results of the uni-modal and fusion models using the time-series input of the PASTIS-R \cite{sainte2021multi} dataset}
\label{tab:fusion_results}
\end{table}

Analysing the results presented in Tab. \ref{tab:fusion_results}, it is clear that fusion methods have been able to achieve superior performance compared to their uni-modal counterparts. Considering the difference between the uni-modal multi-spectral baseline and the proposed Cross Modality Fusion, we observe a more than 5\% increase in RQ and PQ metrics. Furthermore, attention based fusion strategies have achieved a comfortable increase over early and late fusion schemes. Most importantly, the proposed cross attention scheme yields a 0.5\%, 5\%, and 3.8\% performance gain in SQ, RQ, and PQ metrics, respectively, compared to the previous state-of-the-art early fusion mechanism. This denotes the value of extracting salient spatial features across different modalities for informed decision making. By understanding which features are available and where they are important, the overall model is better able to adapt to noisy and occluded observations. 

\subsection{Benchmarking the State-of-the-art Models in an On-board Setting}
\label{sec:benchmarking}

In Tab. \ref{tab:benchmarking_results} we benchmark the baseline models and their attention variants in an on-board setting, where we have used only a single frame from the time-series input. Note that when selecting a frame from the time-series, a random frame is selected and in the multi-modal setting we have used a pair of frames that correspond to the same timestamp from the two modalities. 

\begin{table*}[htbp]
\centering
\begin{tabular}{|c|c|c|c|c|c|c|}
\hline
Model                     & Modality         & SQ & RQ & PQ & Parameters & Runtimes \\ \hline
Detectron 2-FPN R-50 \cite{wu2019detectron2}      & RGB              & 40.98   & 7.43   & 4.4    & 46.0M           &  0.81s        \\ \hline
Detectron 2-FPN R-101 \cite{wu2019detectron2}    & RGB              & 42.14   & 5.87   & 3.6    & 65.0M           & 0.85s         \\ \hline
U-TAE + PAPS  \cite{garnot2021panoptic}            & MutiSpec         & 63.8   & 18.8   & 13.8   & 1.2M           & 0.37s         \\ \hline
U-TAE + PAPS-Late Fusion \cite{sainte2022multi}  & MutiSpec + Radar & 73.4   & 20.5   & 16.1   & 2.3M    & 0.61s          \\ \hline
U-TAE + PAPS-Early Fusion \cite{sainte2022multi} & MutiSpec + Radar & 74.4   & 26.4   & 20.9   & 1.7M    & 0.41s         \\ \hline
Self Attention Fusion     & MutiSpec + Radar & 74.9   & 28.8   & 23.0   & 2.4M           & 0.74s      \\ \hline
Cross Attention Fusion     & MutiSpec + Radar & \textbf{78.1}   & \textbf{31.6}   & \textbf{24.7}   & 2.4M          & 0.81s          \\ \hline
\end{tabular}
\vspace{2px}
\caption{Results of the state-of-the-art models using a single frame input of the PASTIS-R \cite{sainte2021multi} dataset}
\label{tab:benchmarking_results}
\end{table*}

Our analysis reveals the challenge in performing panoptic segmentation from a single frame, without the time-series input. Considering the performance degradation between Tabs. \ref{tab:fusion_results} and \ref{tab:benchmarking_results} for the U-TAE + PAPS models that receives the multispectral inputs, we observe a significant drop of 17.5\%, 30.4\% and 26.6\% in SQ, RQ, and PQ metrics, respectively. We observe that the use of multiple modes helps mitigate the performance drop, however, we also highlight the lack of practicality in adapting such a framework to an on-board evaluation setting given the limited on-board computing power. Therefore, utilising the cross modality attention fusion model that receives a time-series input as our teacher network, we seek to augment the low capacity U-TAE + PAPS network (that receives a single frame from the multi-spectral time-series input).

\subsection{Knowledge Distillation Results}
\label{sec:Knowledge_Distillation_results}

In Tab. \ref{tab:knowledge_distillation_results}, we report the performance of different configurations of student and teacher networks. Note that when choosing the configurations for the student network, we do not consider detectron2, nor do we consider late and early fusion methods for the teacher network, due to the low performance of these networks. Note that for the student network, only the multi-spectral modality of PASTIS-R was provided, while we have considered teacher networks using both uni-modal (multi-spectral input) and multi-modal inputs. In addition, we have used a pre-trained teacher network and trained only the student network to evaluate the off-line knowledge distillation setting. The rest of the teacher networks are jointly trained with the student and thus perform on-line knowledge distillation.

\begin{table*}[htbp]
\centering
\begin{tabular}{|c|c|c|c|c|c|c|}
\hline
Student      & Teacher                   & SQ & RQ & PQ & Parameters (of student) & Run times (of student) \\ \hline
U-Net + PAPS & U-TAE + PAPS (MultiSpec)  & 71.4   & 14.9   & 11.5   &  1.1M                       &  0.32s                     \\ \hline
U-TAE + PAPS & U-TAE + PAPS (MultiSpec)  & 71.6   & 17.0   &  12.7  &  1.2M                       &  0.37s                      \\ \hline
U-Net + PAPS & Self-Attention Fusion           & 70.2   & 23.0   & 17.6   &  1.1M                       &  0.32s                      \\ \hline
U-TAE + PAPS & Self-Attention Fusion           & 73.3   & 27.8   & 21.9   &  1.2M                        &  0.37s                      \\ \hline
U-Net + PAPS & Cross Attention Fusion          & 73.6   & 25.7   & 20.2   &   1.1M                      &  0.32s                     \\ \hline
U-TAE + PAPS & Cross Attention Fusion (off-line) & 74.5   & 27.3   & 21.8   & 1.2M                         & 0.37s                       \\ \hline
U-TAE + PAPS & Cross Attention Fusion          & \textbf{74.5}   & \textbf{30.7}    &  \textbf{24.4}  &  1.2M                        &   0.37s                     \\ \hline
\end{tabular}
\vspace{2px}
\caption{Performance of the knowledge distillation models using a single input frame from the PASTIS-R \cite{sainte2021multi} dataset. Student networks only receive the multi-spectral modality of PASTIS-R. SQ, RQ and PQ are shown for the student network. }
\label{tab:knowledge_distillation_results}
\end{table*}

Analysing the results presented in Tab. \ref{tab:knowledge_distillation_results}, we observe the value of knowledge distillation across all configurations that we evaluate. Specifically, we see a 10.7\%, 11.9\% and 10.6\% increase in  SQ, RQ, and PQ metrics, respectively when the multispectral U-TAE + PAPS model is trained with a Cross Attention Fusion teacher, compared to training the U-TAE + PAPS model alone. Furthermore, we observe that on-line knowledge distillation leads to better knowledge transfer compared to the off-line setting, where the teacher model is pre-trained and frozen. We believe that this is due to the fact that on-line distillation allows the PAPS classifier to better understand how different networks behave in different evaluation conditions and act accordingly, rather than simply trying to re-create the features of the teacher from the student. Moreover, we note a significant performance gap between the U-Net and the U-TAE backbones, resulting from the introduction of L-TAE. Despite the absence of a temporal time-series input, we observe a better encoding through the L-TAE attention scheme. We would like to highlight the fact that $f_{\mathrm{L-TAE}}$ (See eq. \ref{eq:l-tae_att}) is operating on the feature dimension of $e^L$. As such, even though the authors in \cite{garnot2021panoptic} have applied L-TAE to generate temporal attention scores for temporal summarization, we can re-purpose it to better scale the features in the backbone network which results in superior performance compared to the traditional U-Net backbone. 

\subsection{Qualitative Results}
\label{sec:qualitative_results}

\begin{figure*}
    \centering
    \includegraphics[width=.8\textwidth]{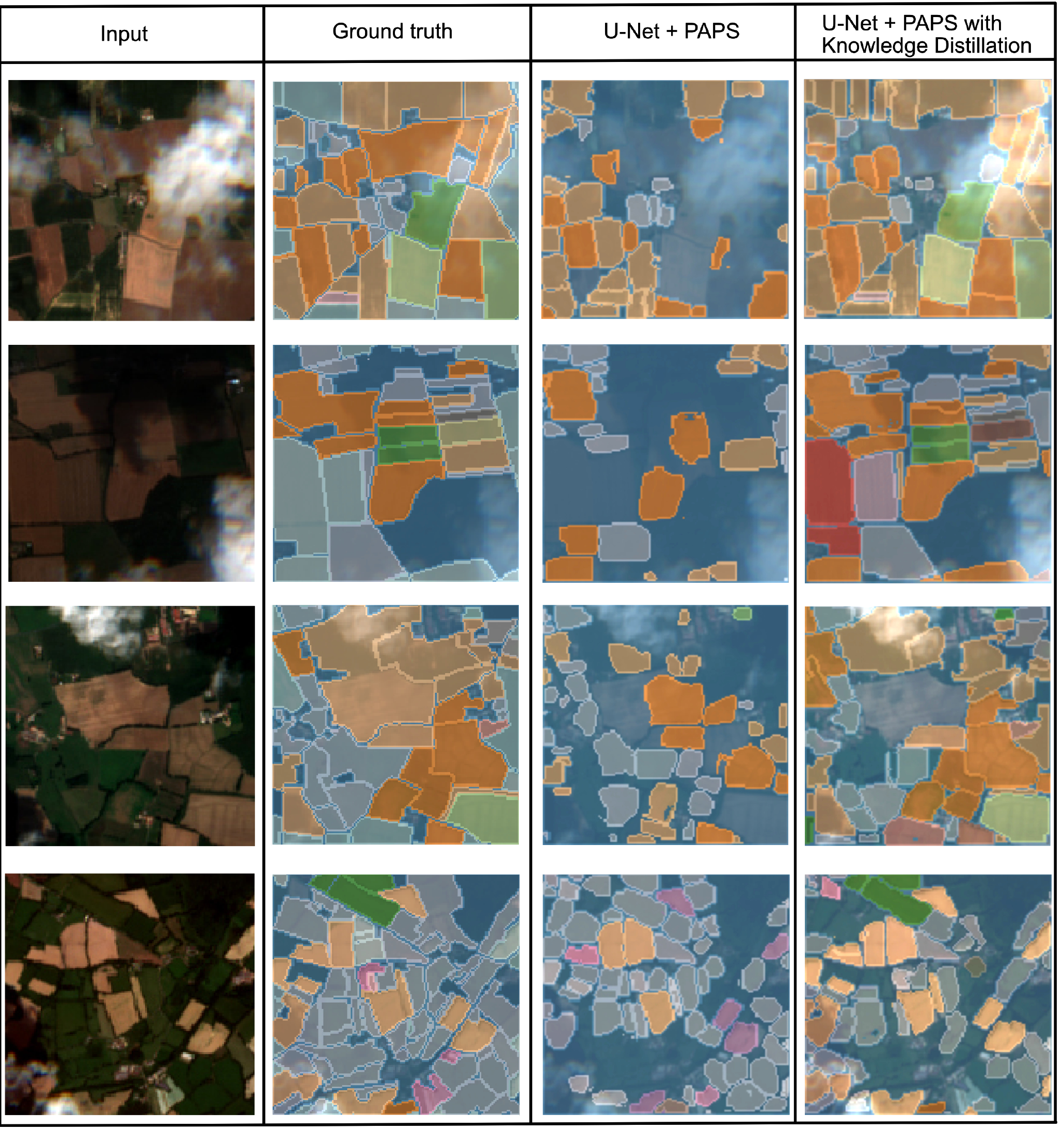}
    \caption{Qualitative results of the baseline U-TAE + PAPS model of \cite{garnot2021panoptic} and the U-TAE + PAPS student network trained with the Cross Att Fusion teacher network using the proposed on-line knowledge distillation framework for a single input frame.}
    \label{fig:qualitative_results}
\end{figure*}

In Fig. \ref{fig:qualitative_results} we present qualitative results when using a single frame from the time-series input of PASTIS-R \cite{sainte2021multi}. The results when the multi-spectral U-TAE + PAPS network is trained with a multi-modal, cross modality attention fusion teacher network are presented together with the baseline U-TAE + PAPS model of \cite{garnot2021panoptic} (trained with a single frame input) for comparisons. 

Figure \ref{fig:qualitative_results} clearly shows that the baseline U-TAE + PAPS network struggles due to cloud cover as well as the illumination inconsistencies and the resultant boundary ambiguities, leading to miss-classifications and missed detection. Correctly identified parcels are significantly fewer compared to the ground truth. Leveraging the time-series input and multi-modality observations our cross modality attention fusion teacher network has been able to positively influence the student network, which allowed the same U-TAE + PAPS student model to generate comparatively better results through the proposed knowledge distillation pipeline. 

\section{Conclusions}
\label{sec:conclusions}
In this article we formulated and evaluated different deep neural network architectures for on-board panoptic segmentation of multi-spectral satellite images. To this end, we first benchmarked existing state-of-the-art panoptic segmentation methodologies in an on-board setting and observed the performance degradation that occurs, largely due to loss of temporal data, which is needed to reduce confusion and ambiguities that arise due to cloud obstruction and other atmospheric noise. We seek to compensate for this loss of information through the introduction of an additional modality, and an augmented multi-modality fusion strategy is proposed using a cross-modality attention fusion scheme. Most importantly, we alleviated additional information processing burden caused by the introduction of multi-modality streams via a knowledge distillation pipeline, where a heavy-weight multi-modality teacher that receives a time-series input was used to guide a light-weight uni-modal student network which only receives a single frame from the multi-spectral time-series. With this approach, a significant 10.7\%, 11.9\% and 10.6\% increase in Segmentation Quality (SQ), Recognition Quality (RQ), and Panoptic Quality (PQ) metrics was observed compared to training the baseline U-TAE + PAPS single-frame uni-modal model alone. In future work, we will investigate methods to further augment the efficiency and the performance of our student model while extending the framework to process hyper-spectral images.

\ifCLASSOPTIONcaptionsoff
  \newpage
\fi



\bibliographystyle{IEEEtran}
\bibliography{egbib}

\begin{thebibliography}{10}
\providecommand{\url}[1]{#1}
\csname url@samestyle\endcsname
\providecommand{\newblock}{\relax}
\providecommand{\bibinfo}[2]{#2}
\providecommand{\BIBentrySTDinterwordspacing}{\spaceskip=0pt\relax}
\providecommand{\BIBentryALTinterwordstretchfactor}{4}
\providecommand{\BIBentryALTinterwordspacing}{\spaceskip=\fontdimen2\font plus
\BIBentryALTinterwordstretchfactor\fontdimen3\font minus
  \fontdimen4\font\relax}
\providecommand{\BIBforeignlanguage}[2]{{%
\expandafter\ifx\csname l@#1\endcsname\relax
\typeout{** WARNING: IEEEtran.bst: No hyphenation pattern has been}%
\typeout{** loaded for the language `#1'. Using the pattern for}%
\typeout{** the default language instead.}%
\else
\language=\csname l@#1\endcsname
\fi
#2}}
\providecommand{\BIBdecl}{\relax}
\BIBdecl

\bibitem{xu2021board}
P.~Xu, Q.~Li, B.~Zhang, F.~Wu, K.~Zhao, X.~Du, C.~Yang, and R.~Zhong,
  ``On-board real-time ship detection in hisea-1 sar images based on cfar and
  lightweight deep learning,'' \emph{Remote Sensing}, vol.~13, no.~10, p. 1995,
  2021.

\bibitem{qiu2021novel}
J.~Qiu, Z.~Zhang, R.~Wang, P.~Wang, H.~Zhang, J.~Du, W.~Wang, Z.~Chen, Y.~Zhou,
  H.~Jia \emph{et~al.}, ``A novel weight generator in real-time processing
  architecture of dbf-sar,'' \emph{IEEE Transactions on Geoscience and Remote
  Sensing}, vol.~60, pp. 1--15, 2021.

\bibitem{smartsat2022}
A.~Sah, Y.~Sun, A.~Bialkowski, K.~Qin, K.~Nguyen, and C.~Fookes, ``Machine
  learning onboard satellites,'' \emph{SmartSat Technical Report}, vol. 010,
  2022.

\bibitem{liu2021novel}
S.~Liu, H.~Zhao, Q.~Du, L.~Bruzzone, A.~Samat, and X.~Tong, ``Novel
  cross-resolution feature-level fusion for joint classification of
  multispectral and panchromatic remote sensing images,'' \emph{IEEE
  Transactions on Geoscience and Remote Sensing}, 2021.

\bibitem{gao2021hyperspectral}
Y.~Gao, W.~Li, M.~Zhang, J.~Wang, W.~Sun, R.~Tao, and Q.~Du, ``Hyperspectral
  and multispectral classification for coastal wetland using depthwise feature
  interaction network,'' \emph{IEEE Transactions on Geoscience and Remote
  Sensing}, 2021.

\bibitem{jung2021boundary}
H.~Jung, H.-S. Choi, and M.~Kang, ``Boundary enhancement semantic segmentation
  for building extraction from remote sensed image,'' \emph{IEEE Transactions
  on Geoscience and Remote Sensing}, 2021.

\bibitem{hong2020multimodal}
D.~Hong, J.~Yao, D.~Meng, Z.~Xu, and J.~Chanussot, ``Multimodal gans: Toward
  crossmodal hyperspectral--multispectral image segmentation,'' \emph{IEEE
  Transactions on Geoscience and Remote Sensing}, vol.~59, no.~6, pp.
  5103--5113, 2020.

\bibitem{kirillov2019panoptic}
A.~Kirillov, K.~He, R.~Girshick, C.~Rother, and P.~Doll{\'a}r, ``Panoptic
  segmentation,'' in \emph{Proceedings of the IEEE/CVF Conference on Computer
  Vision and Pattern Recognition}, 2019, pp. 9404--9413.

\bibitem{de2022panoptic}
O.~L.~F. de~Carvalho, O.~A. de~Carvalho~J{\'u}nior, A.~O. de~Albuquerque, N.~C.
  Santana, D.~L. Borges, R.~A.~T. Gomes, R.~F. Guimar{\~a}es \emph{et~al.},
  ``Panoptic segmentation meets remote sensing,'' \emph{Remote Sensing},
  vol.~14, no.~4, p. 965, 2022.

\bibitem{garnot2021panoptic}
V.~S.~F. Garnot and L.~Landrieu, ``Panoptic segmentation of satellite image
  time series with convolutional temporal attention networks,'' in
  \emph{Proceedings of the IEEE/CVF International Conference on Computer
  Vision}, 2021, pp. 4872--4881.

\bibitem{arechiga2018onboard}
A.~P. Arechiga, A.~J. Michaels, and J.~T. Black, ``Onboard image processing for
  small satellites,'' in \emph{NAECON 2018-IEEE National Aerospace and
  Electronics Conference}.\hskip 1em plus 0.5em minus 0.4em\relax IEEE, 2018,
  pp. 234--240.

\bibitem{gretok2021onboard}
E.~W. Gretok and A.~D. George, ``Onboard multi-scale tile classification for
  satellites and other spacecraft,'' in \emph{2021 IEEE Space Computing
  Conference (SCC)}.\hskip 1em plus 0.5em minus 0.4em\relax IEEE, 2021, pp.
  110--121.

\bibitem{wang2015onboard}
Y.~Wang, S.~Mei, S.~Wan, Y.~Wang, and Y.~Li, ``Onboard image selection for
  small-satellite based remote sensing mission,'' in \emph{2015 IEEE
  International Geoscience and Remote Sensing Symposium (IGARSS)}.\hskip 1em
  plus 0.5em minus 0.4em\relax IEEE, 2015, pp. 1618--1621.

\bibitem{cucchetti2021onboard}
E.~Cucchetti, C.~Latry, G.~Blanchet, J.~Delvit, and M.~Bruno,
  ``Onboard/on-ground image processing chain for high-resolution earth
  observation satellites,'' \emph{The International Archives of Photogrammetry,
  Remote Sensing and Spatial Information Sciences}, vol.~43, pp. 755--762,
  2021.

\bibitem{sophiayati2009onboard}
S.~Sophiayati~Yuhaniz and T.~Vladimirova, ``An onboard automatic change
  detection system for disaster monitoring,'' \emph{International Journal of
  Remote Sensing}, vol.~30, no.~23, pp. 6121--6139, 2009.

\bibitem{yuhaniz2005embedded}
S.~Yuhaniz, T.~Vladimirova, and M.~Sweeting, ``Embedded intelligent imaging
  on-board small satellites,'' in \emph{Asia-Pacific Conference on Advances in
  Computer Systems Architecture}.\hskip 1em plus 0.5em minus 0.4em\relax
  Springer, 2005, pp. 90--103.

\bibitem{bochkovskiy2020yolov4}
A.~Bochkovskiy, C.-Y. Wang, and H.-Y.~M. Liao, ``Yolov4: Optimal speed and
  accuracy of object detection,'' \emph{arXiv preprint arXiv:2004.10934}, 2020.

\bibitem{sandler2018mobilenetv2}
M.~Sandler, A.~Howard, M.~Zhu, A.~Zhmoginov, and L.-C. Chen, ``Mobilenetv2:
  Inverted residuals and linear bottlenecks,'' in \emph{Proceedings of the IEEE
  conference on computer vision and pattern recognition}, 2018, pp. 4510--4520.

\bibitem{deng2009imagenet}
J.~Deng, W.~Dong, R.~Socher, L.-J. Li, K.~Li, and L.~Fei-Fei, ``Imagenet: A
  large-scale hierarchical image database,'' in \emph{2009 IEEE conference on
  computer vision and pattern recognition}.\hskip 1em plus 0.5em minus
  0.4em\relax Ieee, 2009, pp. 248--255.

\bibitem{adams2021hardware}
C.~Adams, J.~Parker, and D.~Cotten, ``A hardware accelerated computer vision
  library for 3d reconstruction onboard small satellites,'' in \emph{2021 IEEE
  Aerospace Conference (50100)}.\hskip 1em plus 0.5em minus 0.4em\relax IEEE,
  2021, pp. 1--14.

\bibitem{goodwill2020adaptively}
J.~Goodwill, D.~Wilson, S.~Sabogal, A.~D. George, and C.~Wilson, ``Adaptively
  lossy image compression for onboard processing,'' in \emph{2020 IEEE
  Aerospace Conference}.\hskip 1em plus 0.5em minus 0.4em\relax IEEE, 2020, pp.
  1--15.

\bibitem{hihara2015onboard}
H.~Hihara, K.~Moritani, M.~Inoue, Y.~Hoshi, A.~Iwasaki, J.~Takada, H.~Inada,
  M.~Suzuki, T.~Seki, S.~Ichikawa \emph{et~al.}, ``Onboard image processing
  system for hyperspectral sensor,'' \emph{Sensors}, vol.~15, no.~10, pp.
  24\,926--24\,944, 2015.

\bibitem{bruhn2020enabling}
F.~C. Bruhn, N.~Tsog, F.~Kunkel, O.~Flordal, and I.~Troxel, ``Enabling
  radiation tolerant heterogeneous gpu-based onboard data processing in
  space,'' \emph{CEAS Space Journal}, vol.~12, no.~4, pp. 551--564, 2020.

\bibitem{gou2021knowledge}
J.~Gou, B.~Yu, S.~J. Maybank, and D.~Tao, ``Knowledge distillation: A survey,''
  \emph{International Journal of Computer Vision}, vol. 129, no.~6, pp.
  1789--1819, 2021.

\bibitem{garcia2018modality}
N.~C. Garcia, P.~Morerio, and V.~Murino, ``Modality distillation with multiple
  stream networks for action recognition,'' in \emph{Proceedings of the
  European Conference on Computer Vision (ECCV)}, 2018, pp. 103--118.

\bibitem{gao2019privileged}
Z.~Gao, J.~Chung, M.~Abdelrazek, S.~Leung, W.~K. Hau, Z.~Xian, H.~Zhang, and
  S.~Li, ``Privileged modality distillation for vessel border detection in
  intracoronary imaging,'' \emph{IEEE transactions on medical imaging},
  vol.~39, no.~5, pp. 1524--1534, 2019.

\bibitem{guo2020online}
Q.~Guo, X.~Wang, Y.~Wu, Z.~Yu, D.~Liang, X.~Hu, and P.~Luo, ``Online knowledge
  distillation via collaborative learning,'' in \emph{Proceedings of the
  IEEE/CVF Conference on Computer Vision and Pattern Recognition}, 2020, pp.
  11\,020--11\,029.

\bibitem{chen2020online}
D.~Chen, J.-P. Mei, C.~Wang, Y.~Feng, and C.~Chen, ``Online knowledge
  distillation with diverse peers,'' in \emph{Proceedings of the AAAI
  Conference on Artificial Intelligence}, vol.~34, no.~04, 2020, pp.
  3430--3437.

\bibitem{sainte2021multi}
V.~Sainte Fare~Garnot, L.~Landrieu, and N.~Chehata, ``Multi-modal temporal
  attention models for crop mapping from satellite time series,'' \emph{arXiv
  e-prints}, pp. arXiv--2112, 2021.

\bibitem{garnot2020lightweight}
V.~S.~F. Garnot and L.~Landrieu, ``Lightweight temporal self-attention for
  classifying satellite images time series,'' in \emph{International Workshop
  on Advanced Analytics and Learning on Temporal Data}.\hskip 1em plus 0.5em
  minus 0.4em\relax Springer, 2020, pp. 171--181.

\bibitem{sainte2022multi}
V.~Sainte Fare~Garnot, L.~Landrieu, and N.~Chehata, ``Multi-modal temporal
  attention models for crop mapping from satellite time series,'' \emph{ISPRS
  Journal of Photogrammetry and Remote Sensing}, vol. 187, pp. 294--305, 2022.

\bibitem{wu2019detectron2}
Y.~Wu, A.~Kirillov, F.~Massa, W.-Y. Lo, and R.~Girshick, ``Detectron2,''
  \url{https://github.com/facebookresearch/detectron2}, 2019.

\end{thebibliography}
%
\

\end{document}